\documentclass{llncs}

\usepackage{epsfig}
\usepackage{graphicx}

\usepackage{todonotes}

\usepackage{enumitem}

\usepackage{amsmath}
\usepackage{amssymb}

\usepackage{url}
\usepackage{hyperref}
\usepackage{xcolor}

\begin{document}

\title{A Decision Tree Approach to Predicting Recidivism in
       Domestic Violence}

\author{Senuri Wijenayake\inst{1} \and Timothy Graham\inst{2} \and
        Peter Christen\inst{2}}
\institute{Faculty of Information Technology,
           University of Moratuwa, Sri Lanka.~ \\
           \email{senuri.wijenayake@gmail.com}
\and
           Research School of Computer Science,
           The Australian National University, \\
           Canberra.~
           \email{\{timothy.graham,peter.christen\}@anu.edu.au}
          }

\maketitle


\begin{abstract}
Domestic violence (DV) is a global social and public health issue
that is highly gendered. Being able to accurately predict DV
recidivism, i.e.,\ re-offending of a previously convicted offender,
can speed up and improve risk assessment procedures for police and
front-line agencies, better protect victims of DV, and potentially
prevent future re-occurrences of DV. Previous work in DV recidivism
has employed different classification techniques, including
decision tree (DT) induction and logistic regression, where the
main focus was on achieving high prediction accuracy. As a result,
even the diagrams of trained DTs were often too difficult to interpret
due to their size and complexity, making decision-making
challenging. Given there is often a trade-off between model accuracy
and interpretability, in this work our aim is to employ DT induction
to obtain both interpretable trees as well as high prediction
accuracy. Specifically, we implement and evaluate different
approaches to deal with class imbalance as well as feature
selection. Compared to previous work in DV recidivism prediction
that employed logistic regression, our approach can achieve
comparable area under the ROC curve results by using only 3 of 11
available features and generating understandable decision trees
that contain only 4 leaf nodes.
\end{abstract}



\section{Introduction}
\label{sec-intro}

Domestic violence (DV), defined as violence, intimidation or abuse
between individuals in a current or former intimate relationship, is
a major social and public health issue. A World Health Organisation
(WHO) literature review across 35 countries revealed that between
10\% and 52\% of women reported at least one instance of physical
abuse by an intimate partner, and between 10\% and 30\% reported
having experienced sexual violence by an intimate
partner~\cite{Kru02}. 

In Australia, evidence shows that one in six women (17\%) and one in
twenty men (6\%) have experienced at least one incidence of DV since
the age of 15~\cite{ABS17,Cox12}, while a recent report found that
one woman a week and one man a month have been killed by their
current or former partner between 2012-13 and 2013-14, and that the
costs of DV are at least \$22 billion per year~\cite{AIH18}. Whilst
DV can affect both partners in a relationship, these statistics
emphasise the highly gendered nature of this problem. More broadly,
gender inequality has been identified as an explaining factor of
violence against women~\cite{Wal14}.

Worryingly, there has been a recent trend of increasing DV in
Australia, particularly in Western Australia (WA) where family
related offences (assault and threatening behaviour) have risen by
32\% between 2014-15 and 2015-16~\cite{WAP18}. Furthermore, police
in New South Wales (NSW) responded to over 58,000 call-outs for
DV related incidents in 2014~\cite{Bul15}, and DV related assault
accounted for about 43\% of crimes against persons in NSW in
2014--15~\cite{NSW15}.

Given this context, besides harm to individuals, DV results in an
enormous cost to public health and the state. Indeed, it is one of
the top ten risk factors contributing to disease burden among adult
women, correlated with a range of physical and mental health
problems~\cite{AIH16,AIH18}.

Despite the importance of this issue, there has been relatively
little research on the risk of family violence and DV offending in
Australia~\cite{Box15,Fit16}. Recent calls have been made to develop
and evaluate risk assessment tools and decision support systems
(DSS) to manage and understand the risk of DV within populations and
to improve targeted interventions that aim to prevent DV and family
violence before it occurs. Whilst risk assessment tools have
attracted criticism in areas such as child protection~\cite{Gil06},
recent studies suggest that DV-related risk assessment tools can be
highly effective in helping police and front-line agencies to make
rapid decisions about detention, bail and victim
assistance~\cite{Mas09,Mes17}.

\smallskip
\textbf{Contributions:}
As discussed in the next section, there are a number of challenges
and opportunities for using data science techniques to improve the
accuracy and interpretability of predictive models in DV risk
assessment. In this paper, we make a contribution to current research
by advancing the use of a DT approach in the context of DV recidivism
to provide predictions about the risk of re-offending that can be
easily interpreted and used for decision-making by front-line
agencies and practitioners. We develop and experimentally evaluate a
technique to reduce the size and complexity of trained DTs whilst
aiming to maintain a high degree of predictive accuracy. 
Our approach is not limited to specialised data collection or single
use cases, but generalises to predicting any DV-related recidivism
using administrative data.

%
%



\section{Related Work}
\label{sec:rel-work}

A key factor when evaluating risk assessment tools and DSS is
determining whether they are able to accurately predict future DV
offending amongst a cohort of individuals under examination. A
standard practice is to measure the accuracy of risk assessment
tools using Receiver Operating Characteristic (ROC) curve
analysis~\cite{Faw04}, which we discuss in more detail later in this
paper. Whilst some tools have been shown to provide reasonably high
levels of predictive accuracy with ROC scores in the high 0.6 to
low 0.7 range~\cite{Ric10}, there are a number of limitations to
current approaches.

In summarising the main limitations with current approaches to
predicting DV offences using risk assessment tools, Fitzgerald and
Graham~\cite{Fit16} argue that such tools ``often rely on detailed
offender and victim information that must form part of specialised
data collection, either through in-take or self-report instruments,
clinical assessment, or police or practitioner observation'' (p. 2).
In this way, the cost in terms of time and money for developing
these tools is prohibitively high and moreover they do not generalise
easily across multiple agencies, social and geographical contexts. 

Although there are presumptions in the literature that the accuracy
and generalisability of such tools may be increased by combining
both official and clinical data sets, studies suggest that the
benefits may be negligible, particularly given the high associated
costs~\cite{Rin13}. Fitzgerald and Graham~\cite{Fit16} posit that
readily available administrative data may be preferable as opposed
to data sets that are more difficult and costly to generate. 
%
They evaluated the potential of existing administrative data drawn
from the NSW Bureau of Crime Statistics and Research (BOCSAR)
Re-offending Database (ROD) to accurately predict violent DV-related
recidivism~\cite{BOC18}. Recidivism is a criminological term that
refers to the rate at which individuals who, after release from
prison, are subsequently re-arrested, re-convicted, or returned to
prison (with or without a new sentence) during a specific time range
following their release~\cite{Ron08}. In this way, a recidivist can
be regarded as someone who is a `repeat' or `chronic' offender.

Fitzgerald and Graham~\cite{Fit16} used logistic regression to
examine the future risk of violent DV offending among a cohort of
individuals convicted of any DV offence (regardless of whether it
is violent or not) over a specific time period. Using ten-fold cross
validation they found the average AUC-ROC (described in
Sect.~\ref{sec:results}) of the models to be $0.69$, indicating a
reasonable level of predictive accuracy on par with other risk
assessment tools described previously. A question that arises from
the study is whether more sophisticated statistical models might be
able to: (1) improve the accuracy for predicting DV recidivism using
administrative data; (2) help to determine and highlight risk
factors associated with DV recidivism using administrative data;
and (3) provide easily interpretable results that can be readily
deployed within risk assessment frameworks. 

A particular approach that has received recent attention is decision
tree (DT) induction~\cite{Qui86}, as we describe in more detail in
the following section. In the context of predicting violent crime
recidivism and in particular DV-related recidivism, Neuilly et
al.~\cite{Neu11} undertook a comparative study of DT induction and
logistic regression and found two main advantages of DTs. First, it
provides outputs that more accurately mimic clinical decisions,
including graphics (i.e., tree drawings) that can be adapted as
questionnaires in decision-making processes. Secondly, the authors
found that DTs had slightly lower error rates of classification
compared to logistic regression~\cite{Neu11}, suggesting that DT
induction might provide higher predictive accuracy compared to
logistic regression. Notably, the related random forest algorithm
has recently been used in DV risk prediction and management, with
reasonably good predictive performance~\cite{Che18} (however, not
considering interpretability which is difficult with random
forests).

While existing work on predicting DV recidivism using logistic 
regression and DT induction is able to obtain results of reasonably 
high accuracy, the important aspect of \emph{interpretability},
i.e.,\  being able to easily understand and explain the prediction
results, has so far not been fully addressed (this is not just the
case in predicting DV recidivism, but also in other areas where data
science techniques are used to predict negative social outcomes
(e.g., disadvantage)~\cite{Wu17}). In our study, described next, we
employ DT induction which will provide both accurate as well as
interpretable results, as we show in our evaluation in
Sect.~\ref{sec:experiments}. 


\section{Decision Tree Based Recidivism Prediction}
\label{sec:approach}

In this study we aim to develop an approach to DV recidivism
prediction that is both accurate and interpretable. We believe
interpretability is as important as high predictive accuracy in a
domain such as crime prediction, because otherwise any prediction
results would not be informative and actionable for users who are
not experts in prediction algorithms (such as criminologists, law
makers, and police forces). We now describe the three major aspects
of our work, DT induction, class balancing, and feature selection, in more detail.

\smallskip
\textbf{Decision tree induction:} Decision tree (DT)
induction~\cite{Lio14} is a supervised classification and prediction
technique with a long history going back over three
decades~\cite{Qui86}. As with any supervised classification method,
a training data set, $\mathbf{D}_R$, is required that contains
ground-truth data, where each record $r = (\mathbf{x},y) \in
\mathbf{D}_R$ consists of a set of input features, $x_i \in
\mathbf{x}$ (with $1 \le i \le m$ and $m=|\mathbf{x}|$ the number
of input features), and a class label $y$. Without loss of
generality we assume $y =\{0,1\}$ (i.e.\ a binary, two-class,
classification problem). The aim of DT induction is, based
on the records in $\mathbf{D}_R$, to build a model in the form of a
tree that is able to accurately represent the characteristics of
the records in $\mathbf{D}_R$. An example DT trained on our DV data
set is shown in Fig.~\ref{fig:example-tree}.

A DT is a data structure which starts with a root node that contains
all records in $\mathbf{D}_R$. Using a heuristic measure such as
information gain or the Gini index~\cite{Lio14}, the basic idea of
DT induction algorithms is to identify the best input feature in
$\mathbf{D}_R$ that splits $\mathbf{D}_R$ into two (or more,
depending upon the actual algorithm used) partitions of highest
purity, where one partition contains those records in
$\mathbf{D}_R$ where most (ideally all) of their class label is
$y=0$ while the other partition contains those records in
$\mathbf{D}_R$ where most (ideally all) of their class label is
$y=1$. This process of splitting is continued recursively until
either all records in a partition are in one class only (i.e., the
partition is pure), or a partition reaches a minimum partition size
(in order to prevent over-fitting~\cite{Lio14}).

At the end of this process, each internal node of a DT corresponds
to a test on a certain input feature, each branch refers to the
outcomes of such a test, and each leaf node is assigned a class
label from $y$ based on the majority of records that are assigned to
this leaf node. For example, in Fig.~\ref{fig:example-tree}, the
upper-most branch classifies records to be in class $y=0$ based on
tests on only two input features (PP and PC, as described in
Table~\ref{tab:variables}).

A trained DT can then be applied on a testing data set,
$\mathbf{D}_S$, where the class labels $y$ of records in
$\mathbf{D}_S$ are unknown or withheld for testing. Based on the
feature values $x_i$ of a test record $r \in \mathbf{D}_S$, a
certain path in the tree is followed until a leaf node is reached.
The class label of the leaf node is then used to classify the test
record $r$ into either class $y=0$ or $y=1$. For detailed algorithms
the interested reader is referred to~\cite{Lio14,Qui86}. As we
describe in more detail in Sect.~\ref{sec:experiments}, we will
explore some parameters for DT induction in order to identify small
trees that are interpretable but achieve high predictive accuracy.

\smallskip
\textbf{Class balancing:}
Many prediction problems in areas such as criminology suffer from a
class imbalance problem, where there is a much smaller number of
training records with class label $y=1$ (e.g., re-offenders) versus
a much larger number of training records with $y=0$ (e.g.,
individuals who do not re-offend). In our DV data set, as described
in detail in Sect.~\ref{sec:experiments}, we have a class imbalance
of around 1:11, i.e., there are 11 times less re-offenders than
those who did not re-offend. Such a high class imbalance can pose
a challenge for many classification algorithms, including
DTs~\cite{Jap02}, because high prediction accuracy can be achieved
by simply classifying all test records as being in the majority
class. From a DV risk prediction perspective, this is highly
problematic because it means that the classifier would predict
every offender as not re-offending~\cite{Che18}. The accuracy would
be high, but such a risk prediction tool would not be useful
in practice.

Two approaches can be employed to overcome this class imbalance
challenge: under-sampling of the majority class and over-sampling
of the minority class~\cite{Dru03}:
\begin{itemize}
\item \emph{Under-sampling of majority class:} Assuming there are
  $n_1 = |\{r = (\mathbf{x},y) \in \mathbf{D}_R: y = 1\}|$ training
  records in class $y=1$ and $n_0 = |\{r = (\mathbf{x},y) \in
  \mathbf{D}_R: y = 0\}|$ training records in class $y=0$, with
  $n_1 + n_0 = |\mathbf{D}_R|$. If $n_1 < n_0$, we can generate a
  balanced training data set by using all training records in
  $\mathbf{D}_R$ where $y=1$, but we sample $n_1$ training records
  from $\mathbf{D}_R$ where $y=0$. As a result we obtain a training
  data set of size $2 \times n_1$ that contains the same number of
  records in each of the two classes $y=0$ and $y=1$.
  \medskip
\item \emph{Over-sampling of minority class:} One potential
  challenge with under-sampling is that the resulting size of the
  training set can become small if the number of minority class
  training records ($n_1$) is small. Under-sampling can also lead to
  a significant loss of detailed characteristics of the majority
  class as only a small fraction of its training records is used for
  training. As a result such an under-sampled training data set
  might not contain enough information to achieve high prediction
  accuracy.
  An alternative is to over-sample the training records from the
  minority class~\cite{Dru03,Jap02}. The basic idea is to replicate
  (duplicate) records from the minority class until the size of the
  minority class ($n_1$) equals the size of the majority class
  ($n_0$), i.e., $n_1 = n_0$.
\end{itemize}
We describe in Sect.~\ref{sec:experiments} how we applied these
two class balancing methods to our DV data set in order to achieve
accurate prediction results.

\smallskip
\textbf{Feature selection:}
Another challenge to interpretable prediction results is the often
increasing number of features in data sets used in many domains.
While having more detailed information about DV offenders, for
example, will likely be useful to improve predictive accuracy, it
potentially can also lead to more complex models such as larger
DTs that are more difficult to employ in practice.

Identifying which available input features are most useful for a
given prediction or classification problem is therefore an
important aspect to obtain interpretable prediction outcomes.
As Hand~\cite{Han06} has shown, the first few most important
features are also those that are often able to achieve almost as
high prediction accuracy as the full set of available features. Any
additional, less predictive feature, included in a model can only
increase prediction accuracy incrementally. There is thus a
trade-off between model complexity, interpretability, and
predictive accuracy. Furthermore, using less features will likely
also result in less time-consuming training times.

Besides interpretability, a second advantage of DTs over other
classification techniques is that the recursive generation of a DT
using the input features available in a training data set is
actually based on a ranking of the importance of the available
features according to a heuristic measure such as information gain
or the Gini index~\cite{Lio14}. The feature with the best value
according to the used measure is the one that is best suited to split
the training data sets into smaller sub-sets of highest purity, as
described above.

Therefore, to identify a ranking of all available input features we
can train a first DT using all available features, and then
remove the least important feature (which has the smallest
information gain or the highest Gini index value~\cite{Lio14}) before
training the next DT, and repeat this process until only one (the
most important) features is left. Assuming a data set contains $m$
input features, we can generate a sequence of $m-1$ DTs that use
from $m$ to only $1$ feature. For each of these trees we calculate
its predictive accuracy and assess its complexity as the size of the
generated tree. Depending upon the requirements of an application
with regard to model complexity (tree size, which affects the
tree's interpretability), and predictive accuracy, a suitable tree
can then be selected.

\smallskip
We illustrate in Algo.~1 our overall approach which incorporates
both class balancing and iterative feature selection. The output of
the algorithm is a list of tuples, each containing a trained DT, the
set of used input features, the size of the tree, and the DT's
predictive quality (calculated as the AUC-ROC and the
F-measure~\cite{Faw04} as described below).
As we discuss in the following section, from the eleven features
available in our data set, not all will be important to predict DV
recidivism. We apply the approach
described in Algo.~1 and investigate both the sizes of the
resulting DTs as well as their predictive accuracy.


\section{Experimental Evaluation}
\label{sec:experiments}

We now describe in more detail the data set we used to evaluate our
DT based prediction approach for recidivism in DV, explain the
experimental setup, and then present and discuss the obtained
results.

  \begin{center}
  \begin{scriptsize}
  \begin{tabular}{ll} \hline\noalign{\smallskip}
\multicolumn{2}{l}{\textbf{Algorithm 1: \emph{Decision tree learning
  with class balancing and feature selection}}} \\
  \noalign{\smallskip} \hline \noalign{\smallskip}
  \multicolumn{2}{l}{Input:} \\
  - $\mathbf{D}_R$: & Training data set \\
  - $\mathbf{D}_S$: & Testing data set \\ 
  - $\mathbf{M}$:   & Set of all input features in $\mathbf{D}_R$
                      and $\mathbf{D}_S$ \\
  - $cb$:           & Class-balancing (sampling) method
    ($under$ or $over$) \\ \noalign{\smallskip}
  \multicolumn{2}{l}{Output:} \\
  - $\mathbf{C}$: & List of classification result tuples \\
    \noalign{\smallskip}
  1:  & $\mathbf{D}_R^0 = \{r = (\mathbf{x},y) \in
         \mathbf{D}_R: y = 0\}$ ~~~~~ // All training records in
         class $y = 0$ \\
  2:  & $\mathbf{D}_R^1 = \{r = (\mathbf{x},y) \in
         \mathbf{D}_R: y = 1\}$  ~~~~~ // All training records in
         class $y = 1$ \\
  3:  & $n_0 = |\mathbf{D}_R^0|$, $n_1 = |\mathbf{D}_R^1|$ ~~~~~~~
        ~~~~~~~~~~ // Number of training records in the two classes \\
  4:  & \textbf{if} $cb = under$ \textbf{then}: \\
  5:  & ~~~ $\mathbf{D}_R^s = \mathbf{D}_R^1 \cup
        sample(\mathbf{D}_R^0, n_1)$ ~~~~~ // Sample $n_1$ training
        records from the majority class \\
  6:  & \textbf{else}: \\
  7:  & ~~~ $\mathbf{D}_R^s = \mathbf{D}_R^0 \cup
         replicate(\mathbf{D}_R^1, n_0)$ ~~~ // Replicate training
         records from minority class \\
  8:  & $\mathbf{C} = []$ ~~~~~~~~~~~~~~~~~~~~~~~~~~~~~~~~~~~~~~~
        // Initialise classification results list \\
  9:  & $\mathbf{M}_u = \mathbf{M}$ ~~~~~~~~~~~~~~~~~~~~~~~~~~~~~~~~~
        ~ // Initialise the set of features to use as all features \\
  10: & \textbf{while} $|\mathbf{M}_u| \ge 1$ \textbf{do}: \\
  11: & ~~~ $\mathbf{dt}_u, lif_u = 
        TrainDecisTree(\mathbf{D}_R^s, \mathbf{M}_u)$ //
        Train tree and get the least important feature \\
  12: & ~~~ $s_u = GetTreeSize(\mathbf{dt}_u)$ \\
  13: & ~~~ $auc_u, fmeas_u = GetPredictionAccuracy(\mathbf{dt}_u,
        \mathbf{D}_S)$ \\
  14: & ~~~ $\mathbf{C}.append([\textbf{dt}_u, \mathbf{M}_u, s_u,
        auc_u, fmeas_u])$ ~~~~ // Append results to results list \\
  15: & ~~~ $\mathbf{M}_u = \mathbf{M}_u \setminus lif_u$ ~~
        ~~~~~~~~~~~~~~~
       // Remove least important feature from current features \\
  16: & \textbf{return} $\mathbf{C}$ \\ \noalign{\smallskip} \hline
  \end{tabular}
 \end{scriptsize}
  \end{center}


\begin{table}[t]
\caption{Independent variables (features) in the ROD data set used
  in the experiments as described in Sect.~\ref{sec:dataset}.
  Variable name abbreviations (in bold) are used in the text.}
  \label{tab:variables}
  \centering
  \begin{scriptsize}
  \begin{tabular}{ll}
  \hline\noalign{\smallskip}
  Variable & Description \\
  \noalign{\smallskip} \hline \noalign{\smallskip}
  \multicolumn{2}{c}{Offender demographic characteristics} \\
    \noalign{\smallskip} \hline \noalign{\smallskip}
  Gender ~(\textbf{G}) & Whether the offender was recorded in ROD as
    male or female. \\
  Age ~(\textbf{A}) & The age category of the offender at the index
    court \\
  ~ & ~~ finalisation was derived from the date of birth of the
    offender \\
  ~ & ~~ and the date of finalisation for the index court
    appearance. \\
  Indigenous status ~(\textbf{IS}) & Recorded in ROD as `Indigenous'
    if the offender had ever \\
  ~ & ~~ identified as being of Aboriginal or Torres
    Strait Islander \\
  ~ & ~~ descent, otherwise `non-Indigenous'. \\
  Disadvantage areas index & Measures disadvantage
    of an offender’s residential postcode at \\
  ~~ (quartiles) ~(\textbf{DA}) & ~~ the index offence. Based on
    the Socio-Economic Index for \\
  ~ & ~~ Areas (SEIFA) score (Australian Bureau of Statistics). \\
    \noalign{\smallskip} \hline \noalign{\smallskip}
  \multicolumn{2}{c}{Index conviction characteristics} \\
    \noalign{\smallskip} \hline \noalign{\smallskip}
  Concurrent offences ~(\textbf{CO}) & Number of concurrent proven
    offences, including the principal \\
  ~ & ~~ offence, at the offender’s index court appearance. \\
  AVO breaches ~(\textbf{AB}) & Number of proven breach of Appended
    Violence Order (AVO) \\
  ~ & ~~ offences at the index court appearance. \\
    \noalign{\smallskip} \hline \noalign{\smallskip}
  \multicolumn{2}{c}{Criminal history characteristics} \\
    \noalign{\smallskip} \hline \noalign{\smallskip}
  Prior juvenile or adult & Number of Youth Justice Conferences or
    finalised court \\
  ~~ convictions ~(\textbf{PC}) & ~~ appearances with any proven
    offence(s) as a juvenile or \\
  ~ & ~~ adult prior to the index court appearance. \\
  Prior serious violent  & Number of Youth Justice Conferences or
    finalised court \\
  ~~ offence conviction &  ~~ appearances in the 5 years prior to
    the reference court \\
  ~~ past 5 years ~(\textbf{P5}) & ~~ appearance with any proven
     homicide or serious assault. \\
  Prior DV-related property & Number of Youth Justice Conferences
    or finalised court \\
  ~~ damage offence conviction & ~~ appearances in the 2 years prior
    to the reference court \\
  ~~ past 2 years ~(\textbf{P2}) & ~~ appearance with any proven DV
     property damage offence. \\
  Prior bonds past 5 years  & Number of finalised court appearances
     within 5 years of the \\
  ~~~ ~(\textbf{PO})~ & ~~ reference court appearance at which
     given a bond. \\
  Prior prison or custodial order~~ & Number of
    previous finalised court appearances at which given \\
  ~~~ ~(\textbf{PP}) & ~~ a full-time prison sentence / custodial
    order. \\
    \noalign{\smallskip} \hline
\end{tabular}
  \end{scriptsize}
\end{table}
    
\textbf{Data set:} \label{sec:dataset}
The data set of administrative data extracted from the NSW Bureau
of Crime Statistics and Research (BOCSAR) Re-offending
Database (ROD)~\cite{BOC18}
consists of $n=14,776$ records, each containing the eleven
independent variables (features) shown in Table~\ref{tab:variables}
as well as the dependent class variable.
The considered features are grouped to represent the offender,
index offence, and criminal history related characteristics of
the offenders.

This study aims to predict whether an offender would re-commit a DV
related offence within a duration of 24 months since the first court
appearance finalisation date (class $y=1$) or not (class $y=0$). DV
related offences in class $y=1$ include any physical, verbal,
emotional, and/or psychological violence or intimidation between
domestic partners.

The Australian and New Zealand Standard Offence Classification
(ANZSOC)~\cite{ANZSOC} has recognised murder, attempted murder and
manslaughter (ANZSOC 111-131), serious assault resulting in injury,
serious assault not resulting in injury and common assault
(ANZSOC 211-213), aggravated sexual assault and non-aggravated
sexual assault (ANZSOC 311-312), abduction and kidnapping and
deprivation of liberty/ false imprisonment (ANZSOC 511-521),
stalking (ANZSOC 291), and harassment and private nuisance and
threatening behaviour (ANZSOC 531-532) as different forms of violent
DV related offences.



\smallskip
\textbf{Experimental Setup:} \label{sec:setup}
As the aim of the study was to provide a more interpretable
prediction to officers involved, a DT classifier along with a
graphical representation of the final decision tree was implemented
using Python version 3.4, where the \emph{scikit-learn}
(\url{http://scikit-learn.org}) machine learning
library~\cite{Ped11} was used for the DT induction (with the Gini
index as feature selection measure~\cite{Lio14}), and tree
visualisations were generated using \emph{scikit-learn} and the
\emph{pydotplus} (\url{https://pypi.python.org/pypi/pydotplus})
package.

In a preliminary analysis we identified that only $8\%$ ($n_1 =
1,182$) of the $14,776$ offenders recommitted a DV offence within
the first 24 months of the finalisation of their index offence.
The data set was thus regarded as imbalanced and we applied the
two class balancing approaches discussed in
Sect.~\ref{sec:approach}: 

\begin{itemize}
\item \emph{Under-sampling of majority class:} The re-offender and
  non-re-offender records were separated into two groups, with
  $1,182$ re-offenders and $13,594$ non-re-offenders respectively.
  Next, we randomly sampled $1,182$ non-re-offender records,
  resulting in a balanced data set containing $2,364$ records.
\item \emph{Over-sampling minority class:} In this approach we
  duplicated re-offender records such that their number ended up to
  be the same as the number of non-re-offender records. The
  resulting balanced data set containing $27,188$ records was then
  shuffled so that the records were randomly distributed.
\end{itemize}

Each of the two balanced data sets were randomly split into a
training and testing set with $70\%$ of all records used to train a
DT and the remaining $30\%$ for testing. We applied the iterative
feature elimination approach described 
in Algo.~1, resulting in a sequence of DTs trained using from 11
and 1 features.

\begin{figure}[t!]
  \centering
  \includegraphics[width=0.44\textwidth]
  {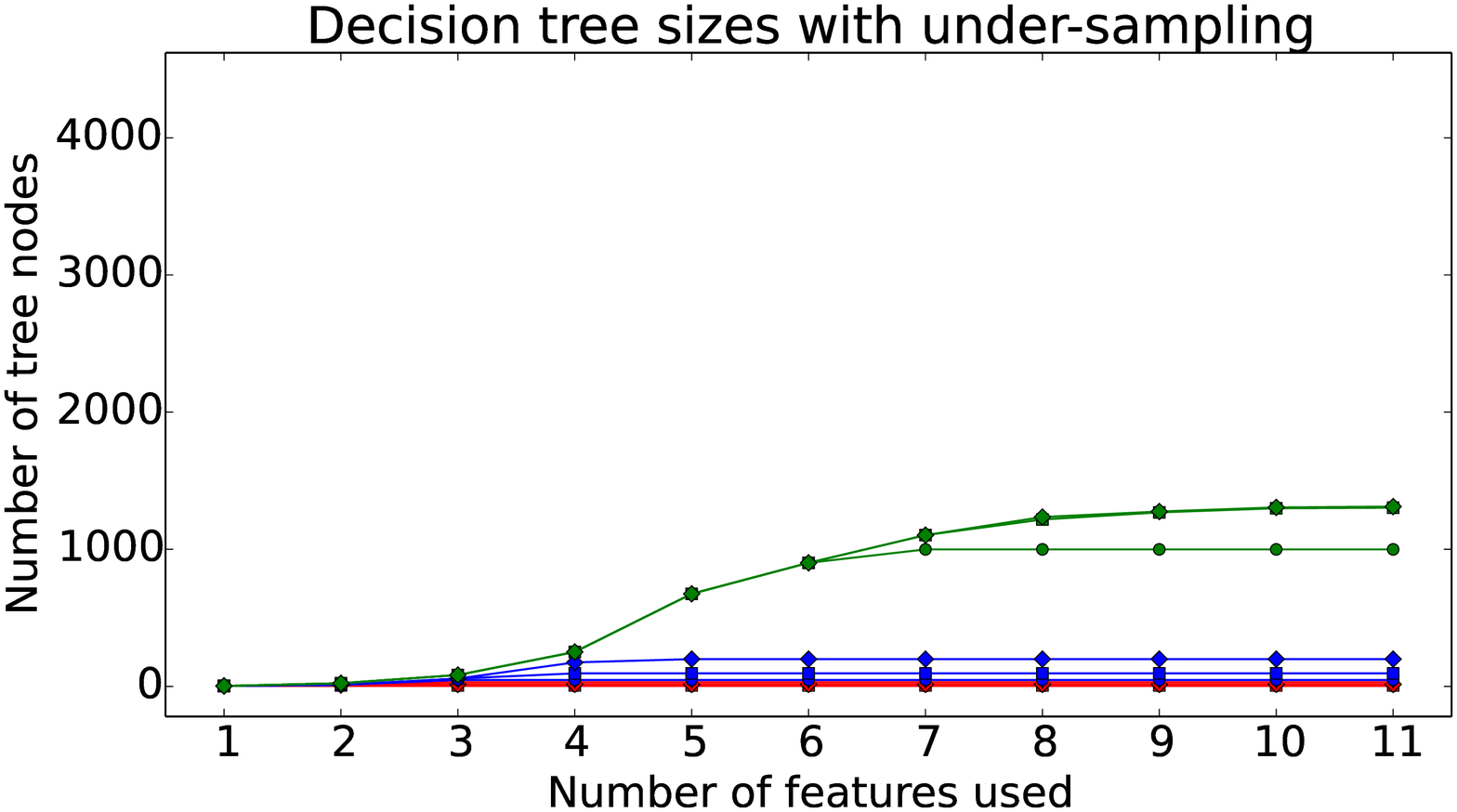}
  ~~~~
  \includegraphics[width=0.44\textwidth]
  {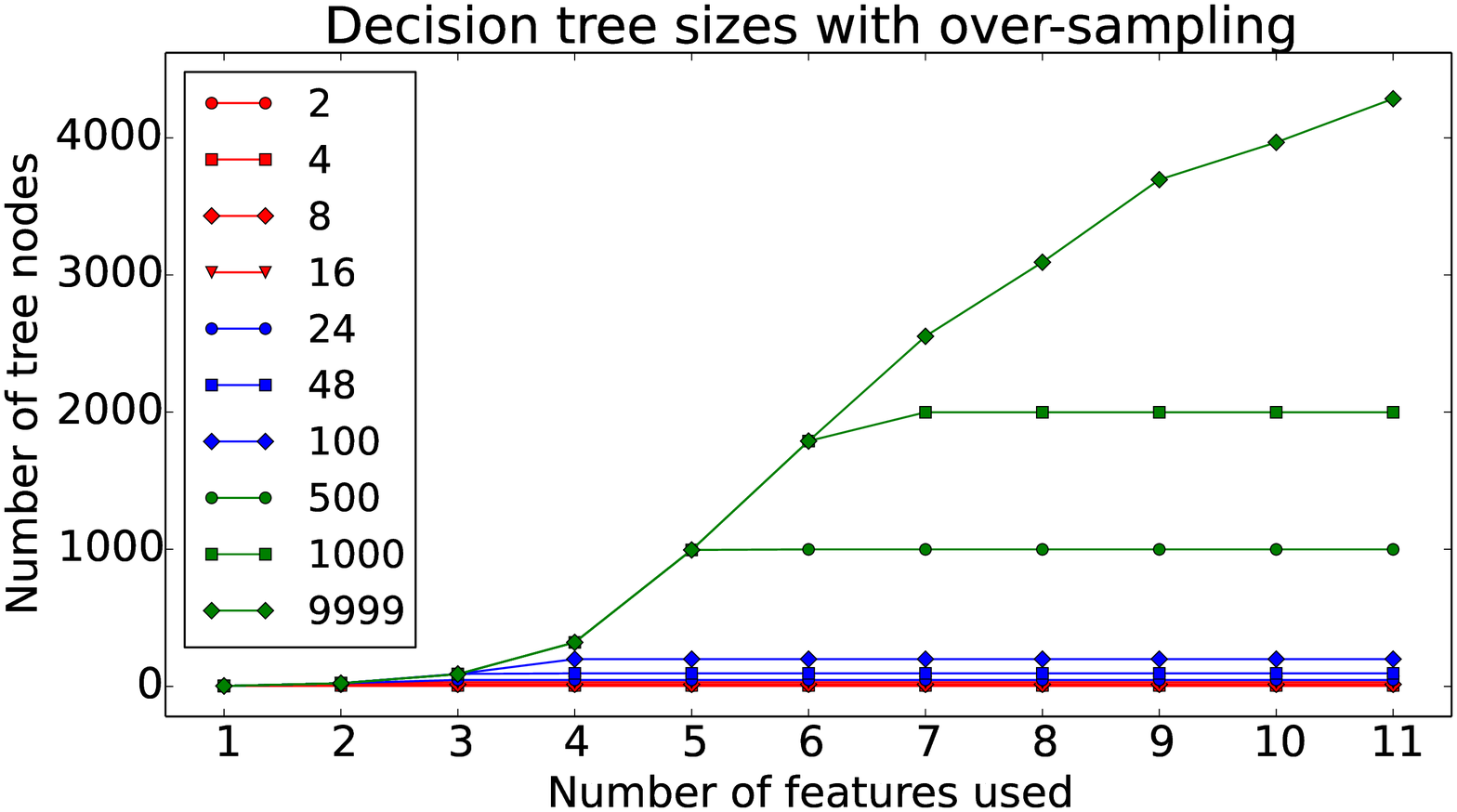}
  ~ \\
  \includegraphics[width=0.44\textwidth]
  {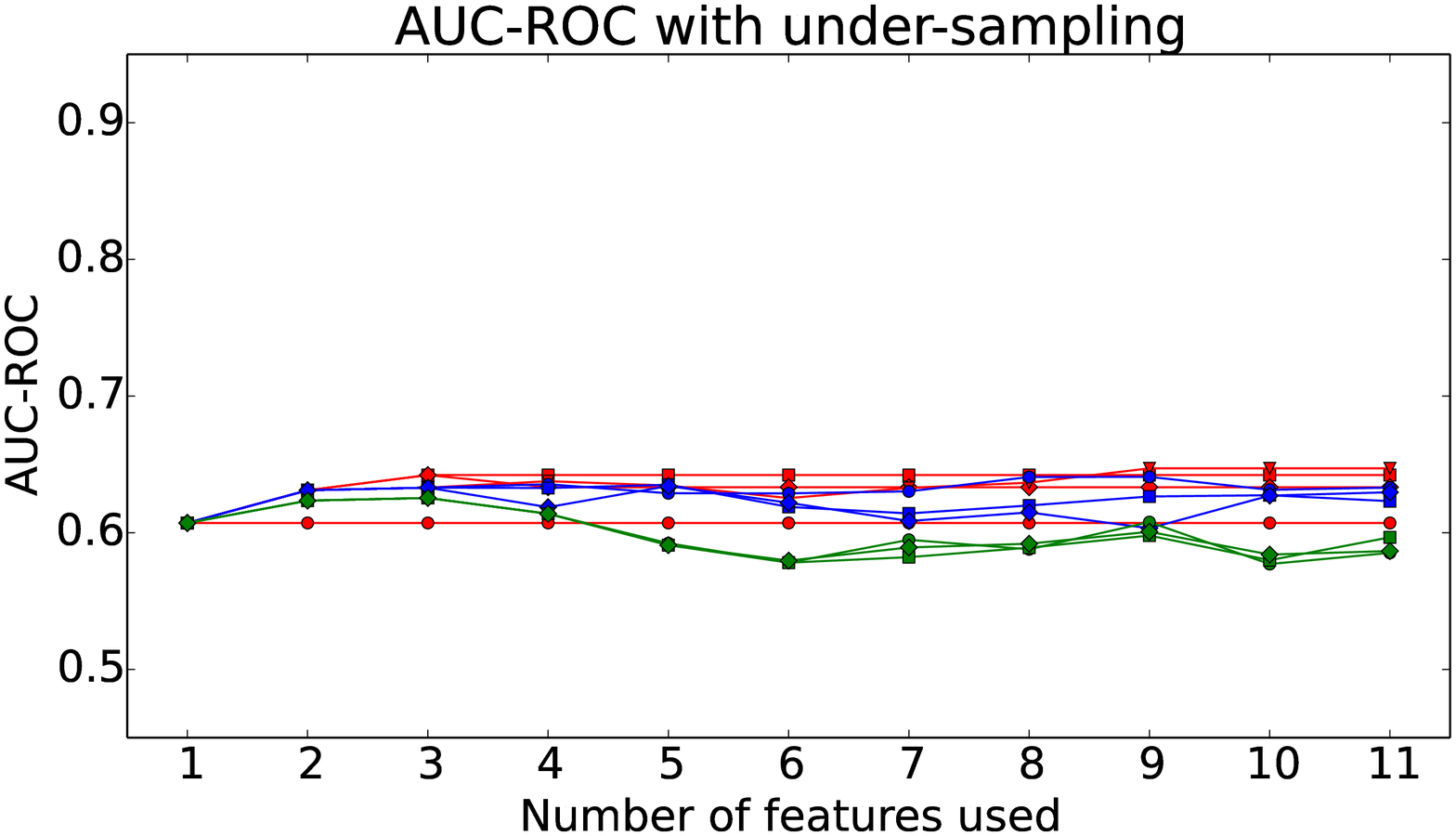}
  ~~~~
  \includegraphics[width=0.44\textwidth]
  {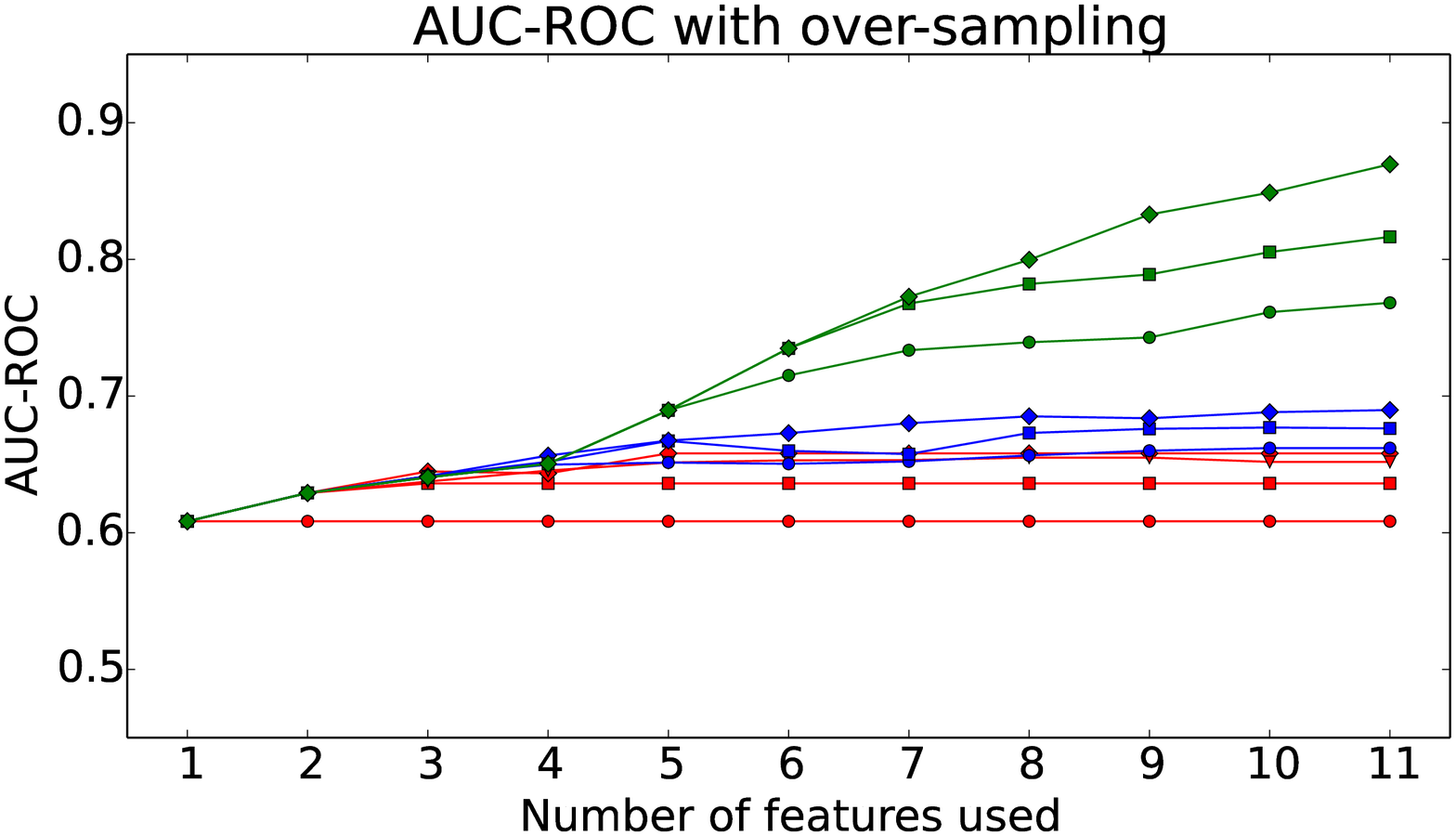}
  ~ \\
  \includegraphics[width=0.44\textwidth]
  {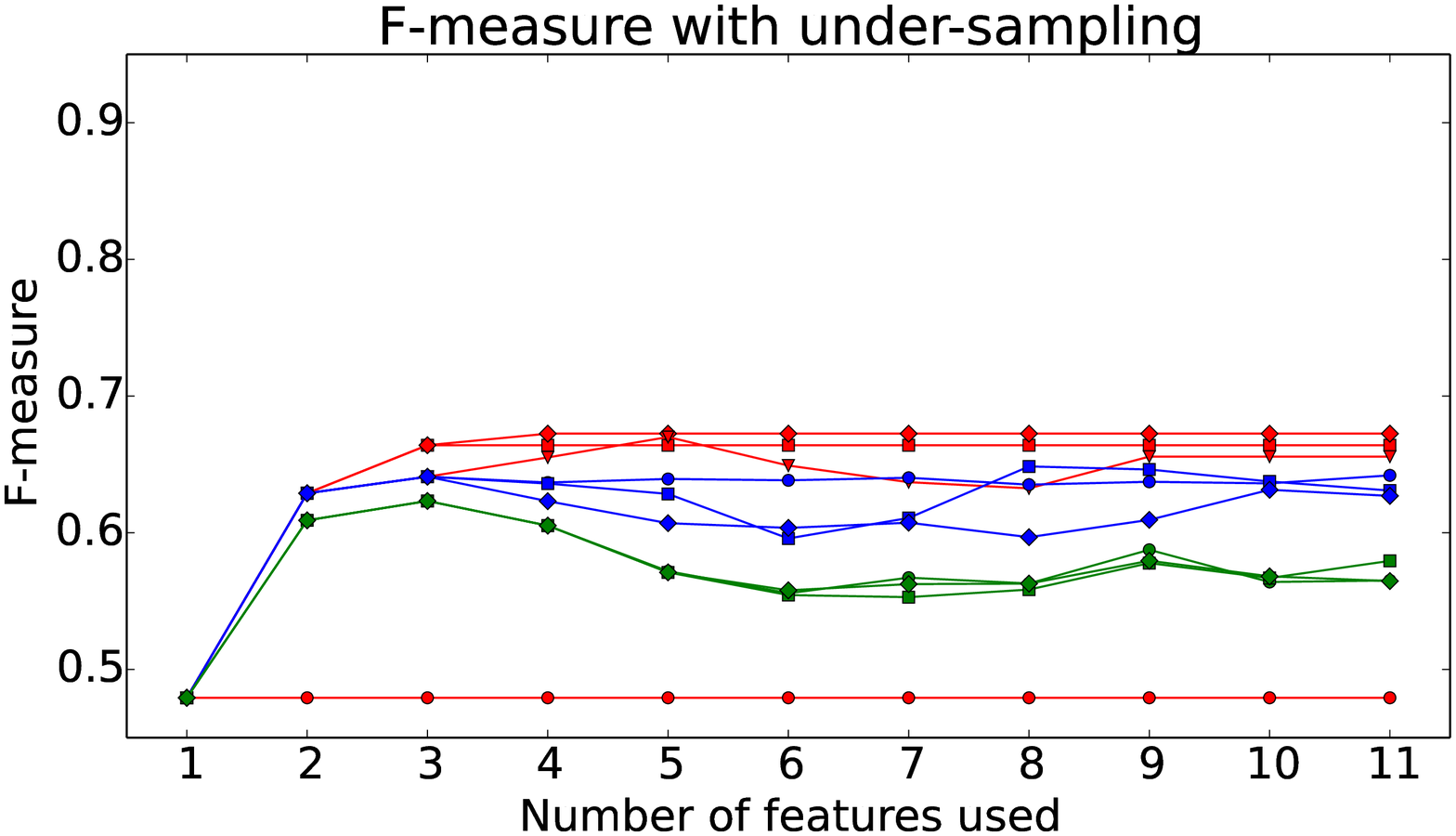}
  ~~~~
  \includegraphics[width=0.44\textwidth]
  {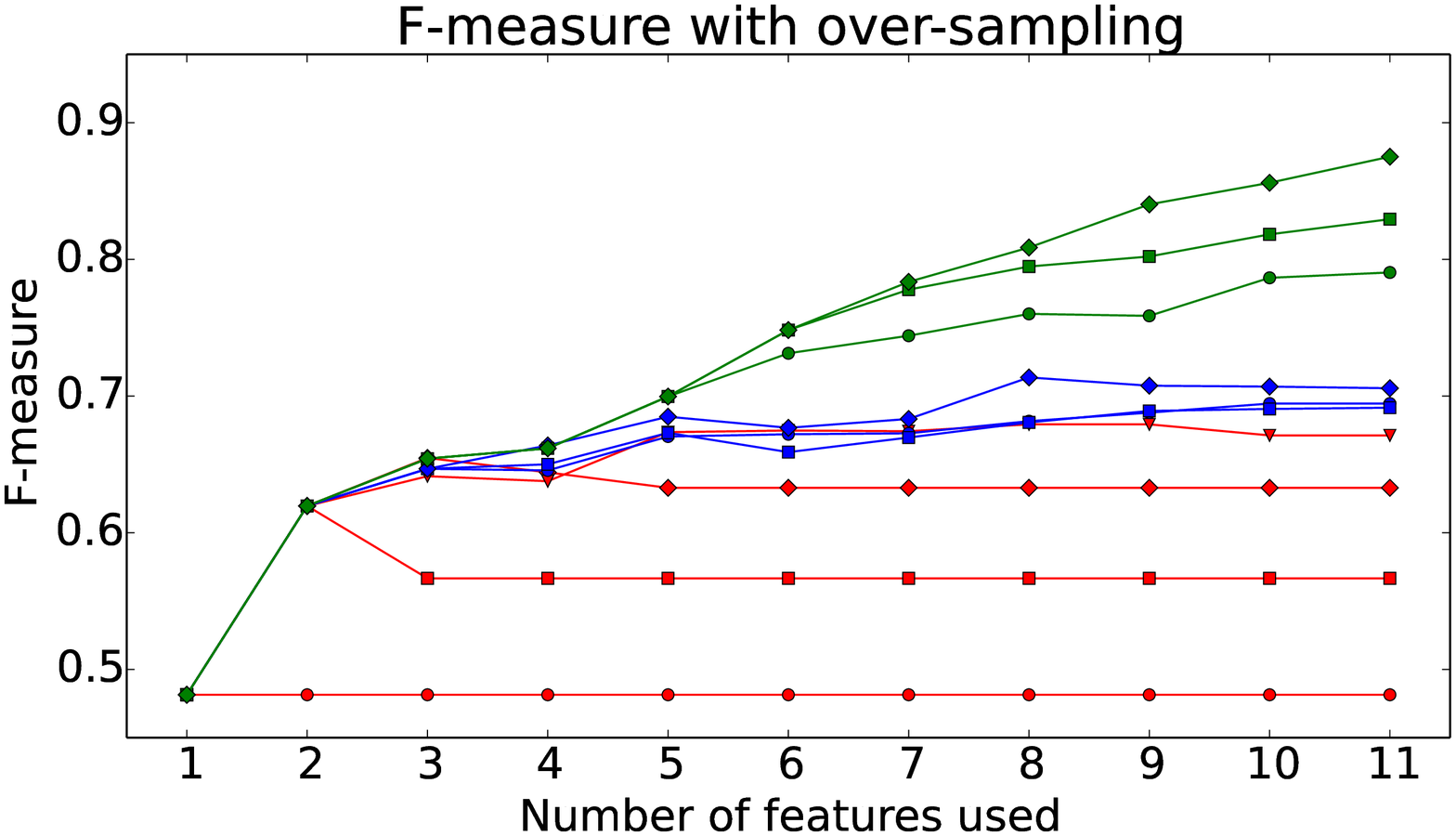}
\caption{Results for different number of features used to train a
         DT: Left with under- and right with over-sampling. The top
         row shows the sizes of the generated DTs, the middle row
         shows AUC-ROC and the bottom row shows F-measure results.
         The parameter varied from $2$ to $9,999$ was the maximum
         number of leaf nodes as described in
         Sect.~\ref{sec:results}.}
         \label{fig:tree-results}
\end{figure}

\begin{figure}[t!]
  \centering
  \includegraphics[width=0.595\textwidth]
  {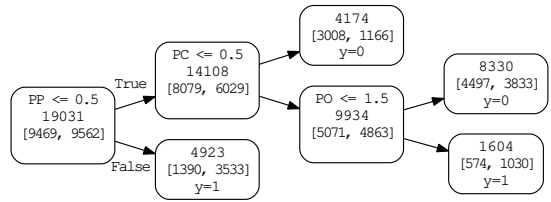}
\caption{A small example decision tree (rotated to the left)
         learned using only three input features (PP, PC, PO, see
         Table~\ref{tab:variables} for descriptions) and
         achieving an AUC-ROC of 0.64. This is almost the same as
         achieved by much larger trees as shown in
         Fig.~\ref{fig:tree-results}, and less than 5\% below a
         previous logistic regression based state-of-the-art
         approach~\cite{Fit16}.}
         \label{fig:example-tree}
\end{figure}

To further explore the ability of DTs of different sizes to obtain
high predictive accuracy, we also varied the \emph{scikit-learn} DT
parameter \emph{max\_leaf\_nodes}, which explicitly stops a DT from
growing once it has reached a certain size. As shown in
Fig.~\ref{fig:tree-results}, we set the value for this parameter
from $2$ (i.e., a single decision on one input feature) to $9,999$
(which basically means no limitation in tree size). While a DT of
limited size might result in reduced predictive accuracy, our aim
was to investigate this accuracy versus interpretability trade-off
which is an important aspect of employing data science techniques
in practical applications such as DV recidivism prediction.

We evaluated the predictive accuracy of the trained DTs using the
commonly used measures of Area Under the Receiver Operator
Characteristic Curve (AUC-ROC), which is calculated as the area
under the curve generated when plotting the true positive rate (TPR)
versus the false positive rate (FPR) at a varying threshold of the
probability that a test record is classified as being in class
$y=1$~\cite{Faw04}. Because the TPR and FPR are always between $0$
and $1$, the resulting AUC-ROC will also be between $0$ and $1$. An
AUC-ROC of $0.5$ corresponds to a random classifier while and
AUC-ROC of $1.0$ corresponds to perfect classification.

As a second measure of predictive accuracy we also calculated the
F-measure \cite{Faw04}, the harmonic mean of precision and recall,
which is commonly used in classification applications. The F-measure
considers and averages both types of errors, namely false positives
(true non-re-offenders wrongly classified as re-offenders) and false
negatives (true re-offenders wrongly classified as
non-re-offenders).


\begin{table}[t]
\caption{Baseline AUC-ROC results as presented in Fitzgerald and
  Graham~\cite{Fit16} using logistic regression on the same data
  set used in our study.}
 \label{tab:baseline}
  \centering
  \begin{scriptsize}
  \begin{tabular}{lcc}
  \hline\noalign{\smallskip}
  Experimental approach & ROC AUC~~ & 95\% Confidence interval \\
    \noalign{\smallskip} \hline \noalign{\smallskip}
  Internal validation (on full data set)~~ & 0.701 & 0.694 -- 0.717 \\
  Ten-fold cross validation & 0.694 & 0.643 -- 0.742 \\
    \noalign{\smallskip} \hline
  \end{tabular}
  \end{scriptsize}
\end{table}







\smallskip
\textbf{Results and Discussion:} \label{sec:results}
In Table~\ref{tab:baseline} we show the baseline results obtained
by a state-of-the-art logistic regression based approach using the
same data set as the one we used. As can be seen, an AUC-ROC of
$0.694$ was obtained, however this approach does not allow easy
interpretation of results due to the logistic regression method
being used.

The detailed results of our DT based approach are shown in
Fig.~\ref{fig:tree-results}, where tree sizes, AUC-ROC and
F-measure results can be seen for different number of input
features used. As can be seen, with large trees (i.e., no tree
growing limits) and using all input features, exceptionally high
prediction results (with AUC-ROC and F-measure of up to $0.9$)
can be achieved. However, the corresponding DTs, which contain over
$4,000$ nodes, will not be interpretable. Additionally, such large
trees would likely overfit the given testing data set.

As can be seen, almost independent of the number of input features
(at least until only around two features were used), a DT can be
trained with an AUC-ROC of around $0.65$, which is less than $5\%$
below the logistic regression based state-of-the-art baseline
approach shown in Table~\ref{tab:baseline}.

As is also clearly visible, the under-sampling approach (resulting
in a much smaller data set than over-sampling) led to worse prediction
accuracy results when using all input features, but also to much
smaller trees. When using only the few most important features the
prediction accuracy of both class balancing methods are very similar.
As can be seen in Table~\ref{tab:ranking}, the overall ranking of
features according to their importance is quite similar, with
criminal history features being prominent in the most important set
of features.

We show one small example DT learned using the over-sampling method
based on only three features in Fig.~\ref{fig:example-tree}. Such a
small tree will clearly be quite easy to interpret by DV experts in
criminology or by police forces.

\section{Conclusion and Future Work}
\label{sec:conclusion}

Domestic Violence (DV) is displaying a rising trend worldwide with a
significant negative impact on the mental and physical health of
individuals and society at large. Having decision support systems
that can assist police and other front-line officers in the
assessment of possible re-offenders is therefore vital.

With regard to predictive tools that can be used by non-technical
users, interpretability of results is as important as high
accuracy. Our study has shown that even small decision trees (DTs),
that are easily interpretable, trained on balanced training data
sets and using only a few input features can achieve predictive
accuracy almost as good as previous state-of-the-art approaches.

As future work, we aim to investigate the problem of producing
interpretable DT models when DV data sets are linked with external
administrative data. This could provide access to additional
features to gain improved insights to the decision making process,
resulting in higher accuracy. While here 
we have used eleven features only, future studies could deploy
hundreds or even thousands of features derived from 
administrative data sources. The experiments conducted in
this study provide a basis to develop methods for maximising both
the accuracy and interpretability of DV risk assessment tools using
Big Data collections.

\begin{table}[t!]
\caption{Feature importance rankings for over- and under-sampling
  as discussed in Sect.~\ref{sec:approach} averaged over all
  parameter settings used in the experiments. The first ranked
  feature is the most important one. Feature codes are from
  Table~\ref{tab:variables}}
 \label{tab:ranking}
  \centering
  \begin{scriptsize}
  \begin{tabular}{lccccccccccc}
  \hline\noalign{\smallskip}
  Sampling & \multicolumn{11}{c}{Feature ranking} \\
  approach & ~ 1~ & ~ 2~ & ~ 3~ & ~ 4~ & ~ 5~ & ~ 6~ & ~ 7~ &
    ~ 8~ & ~ 9~ & ~10~ & ~11~ \\
    \noalign{\smallskip} \hline \noalign{\smallskip}
  Over-sampling & PP & PC & A & DA & PO & CO & IS & P2 & G & AV
    & P5 \\
  Under-sampling~~ & ~PC~ & ~PO~ & ~PP~ & ~CO~ & ~A~ & ~AV~ & ~DA~
    & ~P2~ & ~IS~ & ~G~ & ~P5~ \\
    \noalign{\smallskip} \hline
  \end{tabular}
  \end{scriptsize}
\end{table}

\bibliographystyle{splncs03}

\bibliography{paper}

\begin{thebibliography}{10}
\providecommand{\url}[1]{\texttt{#1}}
\providecommand{\urlprefix}{URL }

\bibitem{ANZSOC}
{Australian and New Zealand Society of Criminology}: http://www.anzsoc.org

\bibitem{ABS17}
{Australian Bureau of Statistics}: Personal safety survey 2016 (2017)

\bibitem{AIH18}
{Australian Institute of Health and Welfare}: Family, domestic and sexual
  violence in {Australia} (2018)

\bibitem{AIH16}
{Australian Institute of Health and Welfare; Australia's National Research
  Organisation for Women's Safety Limited (ANROWS)}: Examination of the health
  outcomes of intimate partner violence against women: State of knowledge paper
  (2016)

\bibitem{Box15}
Boxall, H., Rosevear, L., Payne, J.: Identifying first time family violence
  perpetrators: The usefulness and utility of categorisations based on police
  offence records. Trends and Issues in Crime and Criminal Justice (487) (2015)

\bibitem{Bul15}
Bulmer, C.: Australian police deal with a domestic violence matter every two
  minutes. ABC NEWS (2015)

\bibitem{Cox12}
Cox, P.: Violence against women in {Australia}: {A}dditional analysis of the
  {Australian Bureau of Statistics}' personal safety survey. Horizons Research
  Report, Australia’s National Research Organisation for Women’s Safety
  (ANROWS)  (2012)

\bibitem{Dru03}
Drummond, C., Holte, R.C.: C4. 5, class imbalance, and cost sensitivity: why
  under-sampling beats over-sampling. In: ICML Workshop (2003)

\bibitem{Faw04}
{Fawcett T}: {ROC Graphs}: Notes and practical considerations for researchers.
  Tech. Rep. HPL-2003-4, HP Laboratories, Palo Alto (2004)

\bibitem{Fit16}
Fitzgerald, R., Graham, T.: Assessing the risk of domestic violence recidivism.
  Crime and Justice Bulletin, NSW Bureau of Crime Statistics and Research
  (2016)

\bibitem{Gil06}
Gillingham, P.: Risk assessment in child protection: Problem rather than
  solution? Australian Social Work  59(1),  86--98 (2006)

\bibitem{Han06}
Hand, D.: Classifier technology and the illusion of progress. Statistical
  Science  21(1),  1--14 (2006)

\bibitem{Che18}
Hsieh, T., Wang, Y.H., Hsieh, Y.S., Ke, Jing-Tai, L.C.K., Chen, S.C.: Measuring
  the unmeasurable: A study of domestic violence risk prediction and
  management. Journal of Technology in Human Services  (2018)

\bibitem{Jap02}
Japkowicz, N., Stephen, S.: The class imbalance problem: A systematic study.
  Intelligent data analysis  6(5),  429--449 (2002)

\bibitem{Kru02}
Krug, E., Mercy, J., Dahlberg, L., Zwi, A.: The world report on violence and
  health (2002)

\bibitem{Lio14}
Lior, R., Maimon, O.: Data mining with decision trees: theory and applications,
  vol.~81. World Scientific, 2 edn. (2014)

\bibitem{Mas09}
Mason, R., Julian, R.: Analysis of the {Tasmania Police Risk Assessment
  Screening Tool (RAST)}, final report; {Tasmanian Institute of Law Enforcement
  Studies, University of Tasmania} (2009)

\bibitem{Mes17}
Messing, J.T., Campbell, J., Sullivan~Wilson, J., Brown, S., Patchell, B.: The
  lethality screen: the predictive validity of an intimate partner violence
  risk assessment for use by first responders. Journal of interpersonal
  violence  32(2) (2017)

\bibitem{Neu11}
Neuilly, M.A., Zgoba, K.M., Tita, G.E., Lee, S.S.: Predicting recidivism in
  homicide offenders using classification tree analysis. Homicide studies
  15(2),  154--176 (2011)

\bibitem{NSW15}
{New South Wales Police Force}: {NSW Police Force Annual Report}, 2014-15
  (2015)

\bibitem{BOC18}
{NSW Bureau of Crime Statistics and Research}: Re-offending statistics for
  {NSW} (2018)

\bibitem{Ped11}
Pedregosa, F., Varoquaux, G., Gramfort, A., Michel, V., et~al.: Scikit-learn:
  Machine learning in {Python}. Journal of Machine Learning Research  12(Oct.)
  (2011)

\bibitem{Qui86}
Quinlan, J.R.: Induction of decision trees. Machine learning  1(1),  81--106
  (1986)

\bibitem{Ric10}
Rice, M.E., Harris, G.T., Hilton, N.: Handbook of violence risk assessment,
  chap. The violence risk appraisal guide and sex offender risk appraisal guide
  for violence risk assessment, pp. 99--119. Routledge/Taylor and Francis Group
  (2010)

\bibitem{Rin13}
Ringland, C., et~al.: Measuring recidivism: Police versus court data. BOCSAR
  NSW Crime and Justice Bulletins (175), ~12 (2013)

\bibitem{Ron08}
Ronda, M.: Recividism. In: Encyclopedia of Social Problems, pp. 756--757. Sage
  Publications, Inc (2008)

\bibitem{Wal14}
Wall, L.: Gender equality and violence against women: What's the connection?
  {Australian Institute of Family Studies}, {ACSSA Research Summary}, no.\ 7
  (2014)

\bibitem{WAP18}
{Western Australia Police Force}: Crime in {Western Australia} (2018)

\bibitem{Wu17}
Wu, L., Haynes, M., Smith, A., Chen, T., Li, X.: Generating life course
  trajectory sequences with recurrent neural networks and application to early
  detection of social disadvantage. In: ADMA. pp. 225--242. Springer (2017)

\end{thebibliography}

\end{document}